\def\year{2020}\relax
\documentclass[letterpaper]{article} 
\usepackage{color}
\usepackage{aaai20}  
\usepackage{times}  
\usepackage{helvet} 
\usepackage{courier}  
\usepackage[hyphens]{url}  
\usepackage{graphicx} 
\usepackage{graphicx}  
\usepackage{url}
\usepackage{amsmath}
\usepackage{multirow} 
\usepackage{multicol}
\usepackage{arydshln}
\usepackage{subfigure}
\title{SG-Net: Syntax-Guided Machine Reading Comprehension}
\author{
	Zhuosheng Zhang\textsuperscript{\rm 1,2,3,\thanks{$\ $These authors contribute equally. $\dagger$ Corresponding authors. Part of this work was finished when Zhuosheng Zhang visited NICT. This paper was partially supported by National Key Research and Development Program of China (No. 2017YFB0304100) and Key Projects of National Natural Science Foundation of China (U1836222 and 61733011).}},
	Yuwei Wu\textsuperscript{\rm 1,2,3,4,*},
	Junru Zhou\textsuperscript{\rm 1,2,3},
	Sufeng Duan\textsuperscript{\rm 1,2,3}, \\
	\large \textbf{ 
		Hai Zhao\textsuperscript{\rm 1,2,3,$\dagger$},
		Rui Wang\textsuperscript{\rm 5,$\dagger$}
	}
	\\
	\textsuperscript{\rm 1}Department of Computer Science and Engineering, Shanghai Jiao Tong University\\
	\textsuperscript{\rm 2}Key Laboratory of Shanghai Education Commission for Intelligent Interaction\\
	and Cognitive Engineering, Shanghai Jiao Tong University, Shanghai, China\\
	\textsuperscript{\rm 3}MoE Key Lab of Artificial Intelligence, AI Institute, Shanghai Jiao Tong University, Shanghai, China\\
	\textsuperscript{\rm 4}College of Zhiyuan, Shanghai Jiao Tong University, China\\
	\textsuperscript{\rm 5}National Institute of Information and Communications Technology (NICT), Kyoto, Japan\\
	{\tt\{zhangzs,will8821\}@sjtu.edu.cn, zhaohai@cs.sjtu.edu.cn, wangrui@nict.go.jp} \\
}

\begin{document}

\maketitle

\begin{abstract}
For machine reading comprehension, the capacity of effectively modeling the linguistic knowledge from the detail-riddled and lengthy passages and getting ride of the noises is essential to improve its performance. Traditional attentive models attend to all words without explicit constraint, which results in inaccurate concentration on some dispensable words. In this work, we propose using syntax to guide the text modeling by incorporating explicit syntactic constraints into attention mechanism for better linguistically motivated word representations. In detail, for self-attention network (SAN) sponsored Transformer-based encoder, we introduce syntactic dependency of interest (SDOI) design into the SAN to form an SDOI-SAN with syntax-guided self-attention. Syntax-guided network (SG-Net) is then composed of this extra SDOI-SAN and the SAN from the original Transformer encoder through a dual contextual architecture for better linguistics inspired representation. To verify its effectiveness, the proposed SG-Net is applied to typical pre-trained language model BERT which is right based on a Transformer encoder. Extensive experiments on popular benchmarks including SQuAD 2.0 and RACE show that the proposed SG-Net design helps achieve substantial performance improvement over strong baselines.
\end{abstract}

\section{Introduction}
Recently, much progress has been made in general-purpose language modeling that can be used across a wide range of tasks \cite{radford2018improving,devlin2018bert,zhang2019semantics,zhou2019limit,zhang2019probing}. Understanding the meaning of a sentence is a prerequisite to solve many natural language understanding (NLU) problems, such as machine reading comprehension (MRC) based question answering \cite{Rajpurkar2018Know}. Obviously, it requires a good representation of the meaning of a sentence.

A person reads most words superficially and pays more attention to the key ones during reading and understanding sentences \cite{wang2017learning}. Although a variety of attentive models have been proposed to imitate human learning, most of them, especially global attention methods \cite{bahdanau2014neural} equally tackle each word and attend to all words in a sentence without explicit pruning and prior focus, which would result in inaccurate concentration on some dispensable words \cite{Mudrakarta2018Did}.

We observe that the accuracy of MRC models decreases when answering long questions (shown in Section \ref{sec:long}). Generally, if the text is particularly lengthy and detailed-riddled, it would be quite difficult for deep learning model to understand as it suffers from noise and pays vague attention on the text components, let alone accurately answering questions \cite{zhang2018modeling}. In contrast, existing studies have verified that human reads sentences efficiently by taking a sequence of fixation and saccades after a quick first glance \cite{P17-1172}. 

Besides, for passage involved reading comprehension, a input sequence always consists of multiple sentences. Nearly all of the current attentive methods and language models regard the input sequence as a whole, e.g., a passage, with no consideration of the inner linguistic structure inside each sentence. This would result in process bias caused by much noise and lack of associated spans for each concerned word. 

All these factors motivate us to seek for an informative method that can selectively pick out important words by only considering the related subset of words of syntactic importance inside each input sentence explicitly. With a guidance of syntactic structure clues, the syntax-guided method could give more accurate attentive signals and reduce the impact of the noise brought about by lengthy sentences. 

\begin{figure}
	\centering
	\includegraphics[width=0.45\textwidth]{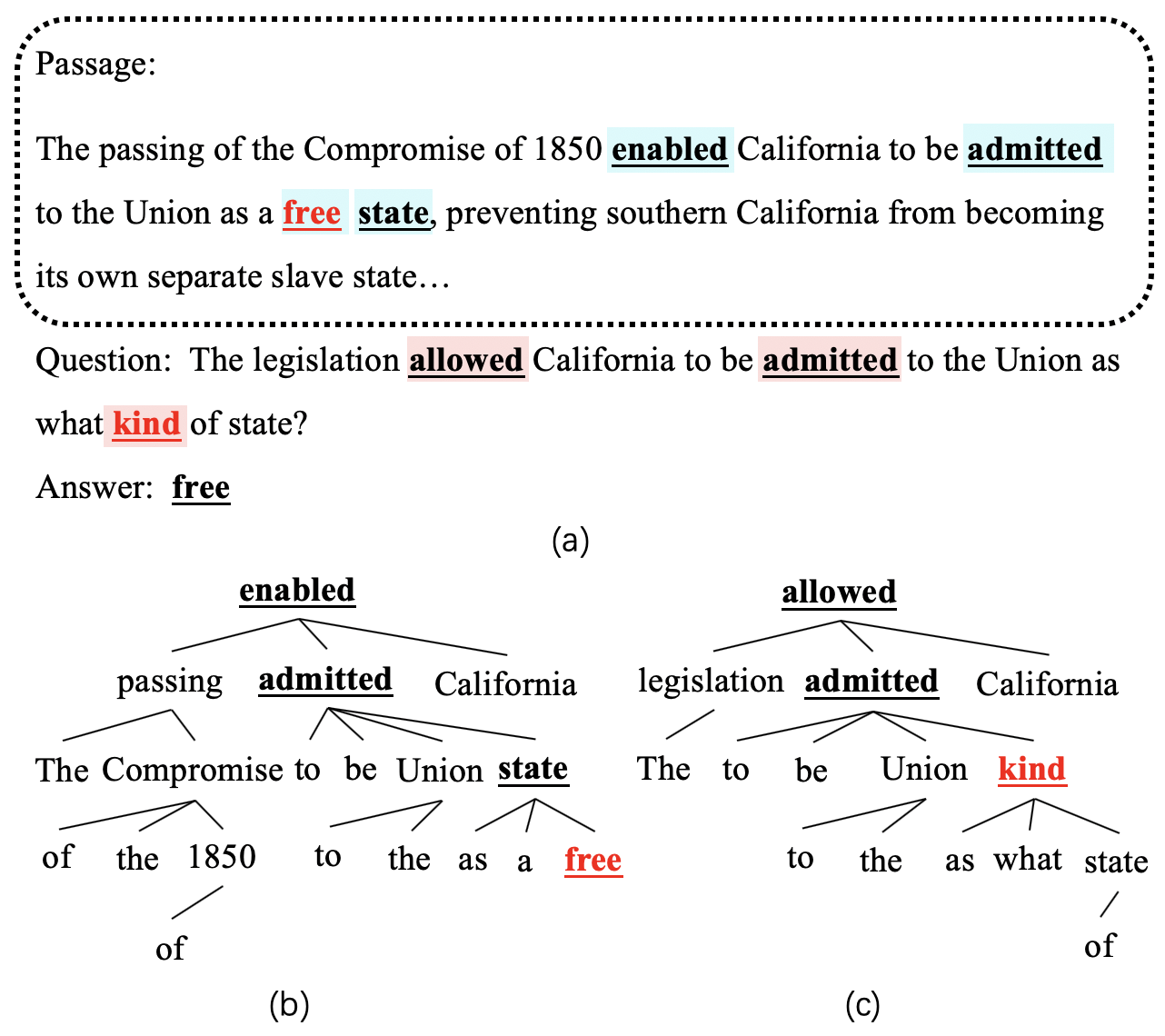}
	\caption{(a) Example of syntax-guided span-based QA. The SDOI of each word (e.g., \emph{free} and \emph{kind}) consists of all its ancestor words and itself marked with the same background color. (b-c) The dependency parsing tree of the given passage sentence and question.}
	\label{exp_dep}
\end{figure}

So far, we have two types of broadly adopted contextualized encoders for building sentence-level representation, RNN-based and Transformer-based \cite{vaswani2017attention}. The latter has shown its superiority which is empowered by a self-attention network (SAN) design. In this paper, we extend the self-attention mechanism with syntax-guided constraint, to capture syntax related parts with each concerned word. Specifically, we adopt pre-trained dependency syntactic parse tree structure to produce the related nodes for each word in a sentence, namely syntactic dependency of interest (SDOI), by regarding each word as a child node and the SDOI consists all its ancestor nodes and itself in the dependency parsing tree. An example is shown in Figure \ref{exp_dep}. 

To effectively accommodate such SDOI information, we propose a novel syntax-guided network (SG-Net), which fuses the original SAN and SDOI-SAN, to provide more linguistically inspired representation for challenging reading comprehension tasks\footnote{Our code is available at \url{https://github.com/cooelf/SG-Net}.}. 

To our best knowledge, we are the first to integrate syntactic relationship as attentive guidance for enhancing state-of-the-art SAN in Transformer encoder. The proposed SG-Net design is applied to pre-trained BERT \cite{devlin2018bert} and evaluated on challenging MRC tasks, which shows its effectiveness by boosting the strong baseline substantially.

\section{Related Work}
\subsection{Machine Reading Comprehension}
In the last decade, the MRC tasks have evolved from the early cloze-style test \cite{hill2015goldilocks,hermann2015teaching} to span-based answer extraction from passage \cite{Rajpurkar2016SQuAD,Nguyen2016MS,Joshi2017TriviaQA,Rajpurkar2018Know} and multi-choice style ones \cite{lai2017race} where the two latter ones are our focus in this work. A wide range of attentive models have been employed, including Attention Sum Reader \cite{kadlec2016text}, Gated attention Reader \cite{Dhingra2017Gated}, Self-matching Network \cite{Wang2017Gated}, Attention over Attention Reader \cite{Cui2017Attention} and Bi-attention Network \cite{Seo2016Bidirectional}.

Recently, deep contextual language model has been shown effective for learning universal language representations by leveraging large amounts of unlabeled data, achieving various state-of-the-art results in a series of NLU benchmarks. Some prominent examples are Embedding from Language models (ELMo), Generative Pre-trained Transformer (OpenAI GPT)  \cite{radford2018improving} and Bidirectional Encoder Representations from Transformers (BERT) \cite{devlin2018bert}. The latest evaluation shows that BERT is powerful and convenient for downstream tasks. Following this line, we extract context-sensitive syntactic features and take pre-trained BERT as our backbone encoder to verify the effectiveness of our proposed SG-Net.

\begin{figure*}
	\centering
	\includegraphics[width=0.8\textwidth]{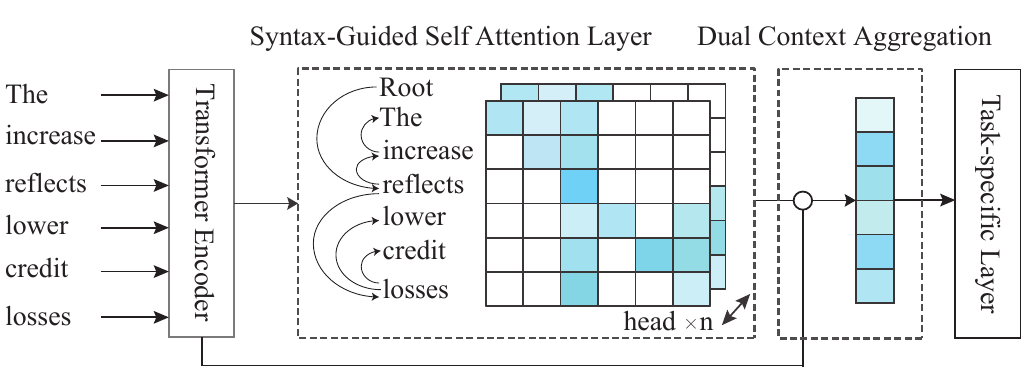}
	\caption{\label{fig:overview} Overview of the syntax-guided network.}
\end{figure*}
\subsection{Syntactic Structures}

Recently, dependency syntactic parsing have been further developed with neural network and attained new state-of-the-art results \cite{zhang-etal-2016-probabilistic,li-etal-2018-seq2seq,Ma2018Stack,li2020global}.
Benefiting from the highly accurate parser, neural network models could enjoy even higher accuracy gains by leveraging syntactic information rather than ignoring it \cite{chen2017neural2,chen2017neural,chen2018syntax,duan2019ialp}. 

Syntactic dependency parse tree provides a form that is capable of indicating the existence and type of linguistic dependency relation among words, which has been shown generally beneficial in various natural language understanding tasks \cite{bowman2016fast}. To effectively exploit syntactic clue, most of previous works \cite{kasai2019syntax} absorb parse tree information by transforming dependency labels into vectors and simply concatenate the label embedding with word representation. However, such simplified and straightforward processing would result in higher dimension of joint word and label embeddings and is too coarse to capture contextual interactions between the associated labels and the mutual connections between labels and words. This inspires us to seek for an attentive way to enrich the contextual representation from the syntactic source. A related work is from \citeauthor{strubell2018linguistically}~(\citeyear{strubell2018linguistically}), which proposed to incorporate syntax with multi-task learning for semantic role labeling. However, their syntax is incorporated by training one extra attention head to attend to syntactic ancestors for each token while we use all the existing heads rather than add an extra one. Besides, this work is based on the remarkable representation capacity of recent language models such as BERT, which have been suggested to be endowed with some syntax to an extent \cite{clark2019does}. Therefore, we are motivated to apply syntactic constraints through syntax guided method to prune the self-attention instead of purely adding dependency features.

In this work, we form a general approach to benefit from syntax-guided representations, which is the first attempt for the SAN architecture improvement in Transformer encoder to our best knowledge. The idea of updating the representation of a word with information from its neighbors in the dependency tree which benefits from explicit syntactic constraints, is well linguistically motivated.

\section{Syntax-Guided Network}
Our goal is to design an effective neural network model which makes use of linguistic information as effectively as possible. We first present the general syntax-guided attentive architecture, building upon the recent advanced Transformer-based encoder\footnote{Note that our method is not limited to cooperate with BERT in our actual use, but any encoder with a self-attention network (SAN) architecture.} and then fit with task-specific layers for machine reading comprehension tasks.

Figure \ref{fig:overview} depicts the whole architecture of our model. Our model first directly takes the output representations from an SAN-empowered Transformer-based encoder, then builds a syntax-guided SAN from the SAN representations. At last, the syntax-enhanced representations are fused from the syntax-guided SAN and the original SAN and passed to task-specific layers for final predictions. 

\subsection{Syntax-Guided Network} \label{SG_ATT}

Our syntax-guided representation is obtained by two steps. Firstly, we pass the encoded representation from the Transformer encoder to a syntax-guided self-attention layer. Secondly, the corresponding output is aggregated with the original encoder output to form a syntax-enhanced representation. It is designed to incorporate the syntactic tree structure information inside a multi-head attention mechanism to indicate the token relationships of each sentence which will be demonstrated as follows.

\paragraph{Syntax-Guided self-attention Layer} 

In this work, we first pre-train a syntactic dependency parser to annotate the dependency structures for every sentence which are then fed to SG-Net as guidance of token-aware attention. Details of the pre-training process of the parser are reported in Section \ref{imp}.

\begin{figure}
	\centering
	\includegraphics[width=3in]{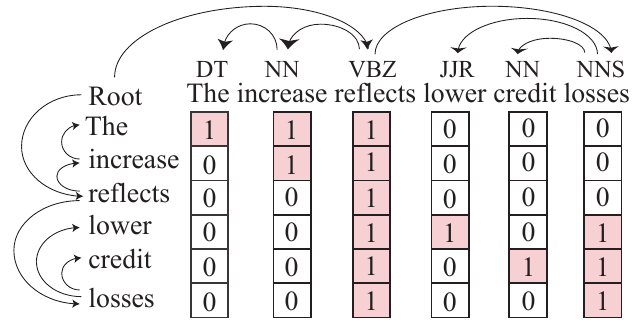}
	\caption{An example of the syntactic dependency of interest (SDOI) mask.}
	\label{dep}
\end{figure}

To use the relationship between head word and dependent words provided by the syntactic dependency tree of sentence, we restrain the scope of attention only between word and all of its ancestor head words\footnote{We extend the idea of using parent in \citeauthor{strubell2018linguistically}~(\citeyear{strubell2018linguistically}) to ancestor for a wider receptive range.}. In other word, we would like to have each word only attend to words of syntactic importance in a sentence, the ancestor head words in the view of the child word. As shown in Figure \ref{dep}, instead of taking attention with each word in whole passage, the word \textit{credit} only makes attention with its ancestor head words \textit{reflects} and \textit{losses} and itself in this sentence, which means that the SDOI of \textit{credit} contains \textit{reflects}, \textit{losses} along with itself\footnote{Note that for special tokens used by BERT such as [CLS], [SEP] and [PAD], the SDOI of these tokens is themselves alone in our implementation, which means these tokens will only attend to themselves in syntax-guided self-attention layer.}.

Specifically, given input token sequence $S=\{s_1,s_2,...,s_n\}$ where $n$ denotes the sequence length, we first use syntactic parser to generate a dependency tree. Then, we derive the ancestor node set $P_i$ for each word $s_i$ according to the dependency tree.  Finally, we learn a sequence of SDOI
mask $\mathcal{M}$, organized as $n*n$ matrix, and elements in each row denote the dependency mask of all words to the row-index word.
\begin{equation}
\mathcal{M}[i,j] = \left\{\begin{matrix}
1, & {\text{ if } j} \in {P_i}\text{ or }j=i \\ 
0, & \text{otherwise}.
\end{matrix}\right.
\end{equation}%
Obviously, if $\mathcal{M}[i,j]=1$, it means that token $s_i$ is the ancestor node of token $s_j$. As the example shown in Figure \ref{dep}, the ancestors of \textit{credit} ($i$=4) are \textit{reflects} ($j$=2), \textit{losses} ($j$=5) along with itself ($j$=4); therefore, $\mathcal{M}[4, (2,4,5)]=1$ and $\mathcal{M}[4, (0,1,3)]=0$.

We then project the last layer output $H$ from the vanilla Transformer into the distinct key, value, and query representations of dimensions $L\times d_k$, $L\times d_q$, and $L\times d_v$, respectively, denoted $K'_i, Q'_i$ and $V'_i$ for each head $i$. Then we perform a dot product to score key-query pairs with the dependency of interest mask to obtain attention weights of dimension $L\times L$, denoted $A_i'$:
\begin{equation}
{A_i}' = \textup{Softmax} \left( {\frac{{\mathcal{M}\cdot \left( {{Q_i}'{K_i}'^T} \right)}}{{\sqrt {{d_k}} }}} \right).
\end{equation}

We then multiply attention weight $A_i'$ by $V_i'$ to obtain the syntax-guided token representations:
\begin{equation}
W_i' = A_i'V_i'.	
\end{equation}
Then $W_i'$ for all heads are concatenated and passed through a feed-forward layer followed by GeLU activations \cite{hendrycks2016bridging}. After passing through another feed-forward layer, we apply a layer normalization to the sum of output and initial representation to obtain the final representation, denoted as $H'=\{h'_1,h'_2,...,h'_n\}$.


\paragraph{Dual Context Aggregation}
 Considering that we have two representations now, one is $H =\{h_1,h_2,...,h_n\}$  from the Transformer encoder, the other is $H'=\{h'_1,h'_2,...,h'_n\}$ from syntax-guided layer from the above part. Formally, the final model output of our SG-Net $\bar H=\{\bar h_1,\bar h_2,...,\bar h_n\}$ is computed by:
\begin{equation}
\bar h_i = \alpha h_i + (1-\alpha)h'_i. \\
\end{equation}

\subsection{Task-specific Adaptation} \label{mrc_models}
We focus on two types of reading comprehension tasks, i.e., \emph{span-based} and \emph{multi-choice} style which can be described as a tuple $<P, Q, A>$ or $<P, Q, C, A>$ respectively, where $P$ is a passage (context) and $Q$ is a query over the contents of $P$, in which a span or choice $C$ is the right answer $A$. For the span-based one, we implemented our model on SQuAD 2.0 task that contains unanswerable questions. Our system is supposed to not only predict the start and end position in the passage $P$ and extract span as answer $A$ but also return a null string when the question is unanswerable. For the multi-choice style, the model is implemented on RACE dataset which is requested to choose the right answer from a set of candidate ones according to given passage and question.

Here, we formulate our model for both of the two tasks and feed the output from the syntax-guided network to task layers according to specific task. Given the passage $P$, the question $Q$, and the choice $C$ specially for RACE, we organize the input $X$ for the encoder as the following two sequences.

\begin{figure}[htb]
	\centering
	\includegraphics[width=0.45\textwidth]{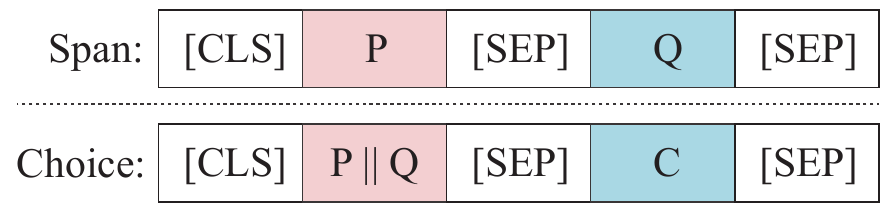}
\end{figure}
\noindent where $||$ denotes concatenation. 

In this work, pre-trained BERT is adopted as our detailed implementation of the Transformer encoder. Thus the sequence is fed to BERT encoder mentioned above to obtain the contextualized representation $\textit{H}$ which is then passed to our proposed syntax-guided self-attention layer and aggregation layer to obtain the final syntax-enhanced representation $\bar{H}$. To keep simplicity, the downstream task-specific layer basically follows the implementation of BERT. We outline below to keep the integrity of our model architecture. For span-based task, we feed $\bar{H}$ to a linear layer and obtain the probability distributions over the start and end positions through a softmax. For multi-choice task, we feed it into the classifier to predict the choice label for the multi-choice model.

\paragraph{SQuAD 2.0}
For SQuAD 2.0, our aim is a span of answer text, thus we employ a linear layer with SoftMax operation and feed $\bar H$ as the input to obtain the start and end probabilities, $s$ and $e$:
\begin{equation}
s, e = \textup{SoftMax}(\textup{Linear}(\bar H)).
\end{equation}


The training objective of our SQuAD model is defined as cross entropy loss for the start and end predictions, 
\begin{equation}
\begin{split}
\mathcal{L}_{has} = y_s \log{s} + y_e \log{e}.\\
\end{split}
\end{equation}
For prediction, given output start and end probabilities $s$ and $e$, we calculate the has-answer score $score_{has}$ and the no-answer score $score_{na}$:
\begin{equation}
\begin{split}
score_{has} & =\max (s_k + e_l),0\le k \le l \le n, \\
score_{na} &= s_0+e_0.
\end{split}
\end{equation}

We obtain a difference score between has-answer score and the no-answer score as final score. A threshold $\delta$ is set to determine whether the question is answerable, which is heuristically computed in linear time with dynamic programming according to the development set. The model predicts the answer span that gives the has-answer score if the final score is above the threshold, and null string otherwise.

\paragraph{RACE}
As discussed in \citeauthor{devlin2018bert}~(\citeyear{devlin2018bert}), the pooled representation explicitly includes classification information during the pre-training stage of BERT. We expect the pooled to be overall representation of the input. Thus, the first token representation $\bar h_0$ in $\bar H$ is picked out and is passed to a feed-forward layer to give the prediction $p$. For each instance with $n$ choice candidates, we update model parameters according to cross-entropy loss during training and choose the one with highest probability as the prediction when testing. 
The training objectives of our RACE model is defined as, $L(\theta) = -\frac{1}{N}\sum_i y_i \log{p}_i$, where $p_{i}$ denotes the prediction, $y_{i}$ is the target, and $i$ denotes the data index.

\section{Experiments}
\subsection{Dataset and Setup}
Our experiments and analysis are carried on two data sets, involving span-based and multi-choice MRC and we use the fine-tuned cased BERT (whole word masking) as the baseline. 
\paragraph{Span-based MRC} As a widely used MRC benchmark dataset, SQuAD 2.0  \cite{Rajpurkar2018Know} combines the 100,000 questions in SQuAD 1.1 \cite{Rajpurkar2016SQuAD} with over 50,000 new, unanswerable questions that are written adversarially by crowdworkers to look similar to answerable ones. For the SQuAD 2.0 challenge, systems must not only answer questions when possible, but also abstain from answering when no answer is supported by the paragraph. Two official metrics are selected to evaluate the model performance: Exact Match (EM) and a softer metric F1 score, which measure the weighted average of the precision and recall rate at a character level.

\paragraph{Multi-choice MRC}
Our multi-choice MRC is evaluated on Large-scale ReAding Comprehension Dataset From Examinations (RACE) dataset \cite{lai2017race}, which consists of two subsets: RACE-M and RACE-H corresponding to middle school and high school difficulty levels. RACE contains 27,933 passages and 97,687 questions in total, which is recognized as one of the largest and most difficult datasets in multi-choice MRC. The official evaluation metric is accuracy.

\begin{figure*}
	\centering
	\includegraphics[width=0.95\textwidth]{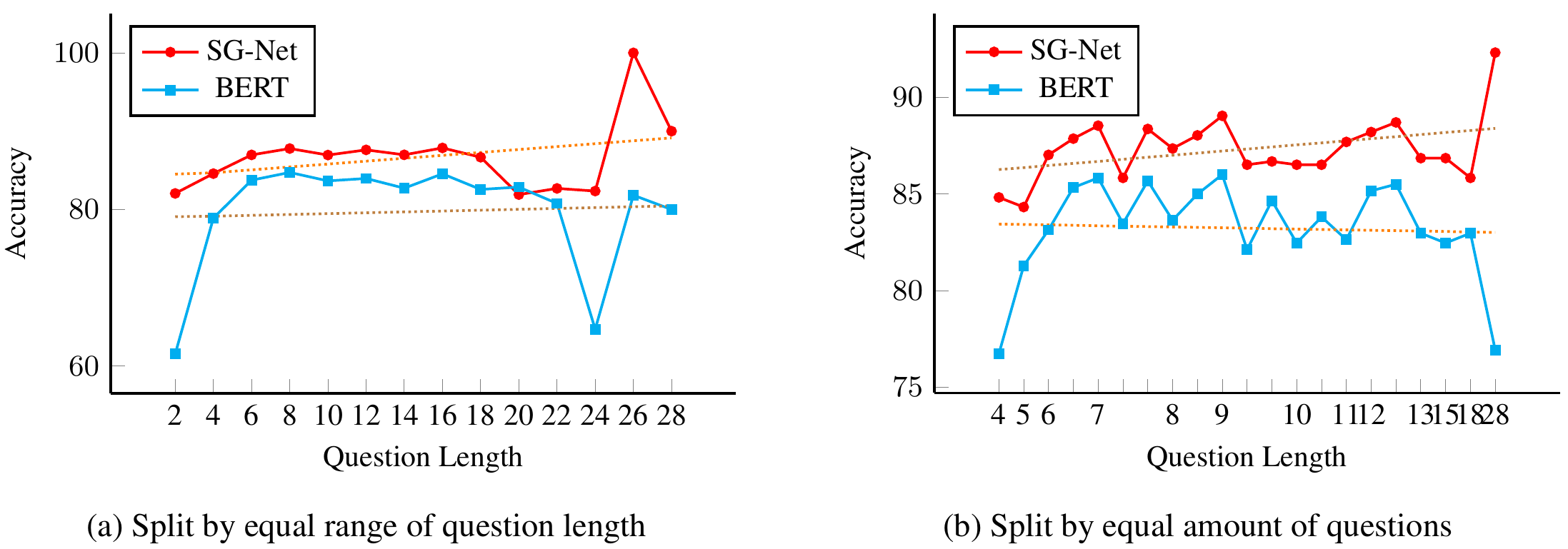}
	\caption{\label{fig:length} Accuracy for different question length. Each data point means the accuracy for the questions in the same length range (a) or of the same number (b) and the horizontal axis in (b) shows that most of questions are of length 7-8 and 9-10.}
\end{figure*}

\subsection{Implementation} 
\label{imp}
For the syntactic parser, we adopt the dependency parser from \citeauthor{zhou2019head}~(\citeyear{zhou2019head}) by joint learning of  constituent parsing \cite{Kitaev-2018-SelfAttentive} using BERT as sole input which achieves very high accuracy: 97.00\% UAS and 95.43\% LAS on the English dataset Penn Treebank (PTB) \cite{MarcusJ93-2004} test set\footnote{We report the results without punctuation of the labeled and unlabeled attachment scores (LAS, UAS).}. Note this work is done in data preprocessing and our parser is not updated with the following MRC models.

For MRC model implementation, We adopt the Whole Word Masking BERT as the baseline \footnote{It is further improved as strong baseline by synthetic self training following \url{https://nlp.stanford.edu/seminar/details/jdevlin.pdf}.}. The initial learning rate is set in \{8e-6, 1e-5, 2e-5, 3e-5\} with warm-up rate of 0.1 and L2 weight decay of 0.01. The batch size is selected in \{16, 20, 32\}. The maximum number of epochs is set to 3 or 10 depending on tasks. The weight $\alpha$ in the dual context aggregation is 0.5. All the texts are tokenized using wordpieces, and the maximum input length is set to 384 for both of SQuAD and RACE. The configuration for multi-head self-attention is the same as that for BERT. 

\begin{table}
	\centering
	\resizebox{\linewidth}{!}
	{
		\begin{tabular}{l c c c c }
			\hline
			\hline
			\multirow{2}{*}{\textbf{Model} }& \multicolumn{2}{c}{\textbf{Dev}} & \multicolumn{2}{c}{\textbf{Test}}\\
			& \textbf{EM} & \textbf{F1}&   \textbf{EM} & \textbf{F1}\\
			\hline
			\multicolumn{5}{c}{\emph{Regular Track}} \\
			Joint SAN &69.3 & 72.2 & 68.7 & 71.4\\	
			U-Net & 70.3  & 74.0  & 69.2 & 72.6 \\	
			RMR + ELMo + Verifier & 72.3  & 74.8 & 71.7 & 74.2 \\
			\hline
			\multicolumn{5}{c}{\emph{BERT Track}} \\
			Human& -  & - & 86.8&	89.5 \\
			\hdashline
			BERT + DAE + AoA$\dagger$& -  & -   & 85.9 & 88.6 \\
			BERT + NGM + SST$\dagger$& -  & - & 85.2 & 87.7\\
			BERT + CLSTM + MTL + V$\dagger$& -  & -   & 84.9 &	88.2 \\
			SemBERT$\dagger$& -  & -  & 84.8 &	87.9   \\
			Insight-baseline-BERT$\dagger$& -  & - & 84.8 &	87.6 \\
			BERT + MMFT + ADA$\dagger$& -  & -  & 83.0 & 85.9 \\
			BERT$_\text{LARGE}$ & -  & - &  82.1 &  84.8 \\
			\hline
			Baseline & 84.1  & 86.8 & -  & - \\
			\textbf{SG-Net}  & \textbf{85.1} & \textbf{87.9} & - & -\\
			\textbf{+Verifier} &  \textbf{85.6} & \textbf{88.3}  & \textbf{85.2} & \textbf{87.9}  \\

			\hline
			\hline
		\end{tabular}
	}
	\caption{\label{tab:squad2.0} Exact Match (EM) and F1 scores (\%) on SQuAD 2.0 dataset for single models. Our model is in boldface. $\dagger$ refers to unpublished work. Besides published works, we also list competing systems on the SQuAD leaderboard at the time of submitting SG-Net (May 14, 2019). Our model is significantly better than the baseline BERT with p-value $<$ 0.01.}
\end{table}
\subsection{Main Results}

To focus on the evaluation of syntactic advance and keep simplicity, we only compare with single models instead of ensemble ones.

\paragraph{SQuAD 2.0}

Table \ref{tab:squad2.0} shows the result on SQuAD 2.0. Various state of the art models from the official leaderboard are also listed for reference. We can see that the performance of BERT is very strong. However, our model is more powerful, boosting the BERT baseline essentially. It also outperforms all the published works and achieves the 2nd place on the leaderboard when submitting SG-NET. We also find that adding an extra answer verifier module could yield better result, which is pre-trained only to determine whether question is answerable or not with the same training data as SG-Net. The logits of the verifier are weighted with $score_{na}$ to give the final predictions.

\begin{table}
	\centering
	{
		\begin{tabular}{l c c c}
			\hline
			\hline
			\textbf{Model} &   \textbf{RACE-M}  & \textbf{RACE-H} & \textbf{RACE}	\\
			\hline
			\multicolumn{4}{c}{\emph{Human Performance}} \\
			Turkers  & 85.1 & 69.4 & 73.3\\
			Ceiling  & 95.4 & 94.2& 94.5 \\
			\hline
			\multicolumn{4}{c}{\emph{Leaderboard}} \\
			DCMN  & 77.6 & 70.1  & 72.3\\
			BERT$_\text{LARGE}$ 	& 	76.6 & 70.1 & 72.0  \\
			OCN  & 76.7 & 69.6 & 71.7  \\ 
			\hline
			Baseline & 78.4 & 70.4& 72.6 \\
			\textbf{SG-Net} & \textbf{78.8} & \textbf{72.2} & \textbf{74.2}\\
			\hline
			\hline
		\end{tabular}
	}
	\caption{\label{tab:race} Accuracy (\%) on RACE test set for single models. Our model is significantly better than the baseline BERT with p-value $<$ 0.01.}
\end{table}

\paragraph{RACE} For RACE, we compare our model with the following latest baselines: Dual Co-Matching Network (DCMN) \cite{zhang2019dual}, Option Comparison Network (OCN) \cite{ran2019option}, Reading Strategies Model (RSM) \cite{sun2018improving}, and Generative Pre-Training (GPT) \cite{radford2018improving}. Table \ref{tab:race} shows the result\footnote{Our concatenation order of $P$ and $Q$ is slightly different from the original BERT. Therefore, the result of our BERT baseline is higher than the public one on the leaderboard, thus our improved BERT implementation is used as the stronger baseline for our evaluation.}. Turkers is the performance of Amazon Turkers on a random subset of the RACE test set. Ceiling is the percentage of unambiguous questions in the test set. From the comparison, we can observe that our model outperforms all baselines, which verifies the effectiveness of our proposed syntax enhancement.

\section{Discussions}

\subsection{Effect of Answering Long Questions}
\label{sec:long}
We sort the questions from SQuAD dev set according to the length and group them into 20 subsets split by equal range of question length and equal amount of questions\footnote{Since the question length is at variance, we depict the two aspects to show the discovery comprehensively.}. Then we calculate the exact match accuracy of the baseline and SG-Net per group, as shown in Figure \ref{fig:length}. We observe that the performance of the baseline drops heavily when encountered with long questions, especially for those longer than 20 words while our proposed SG-Net works robustly, even showing positive correlation between accuracy and length. This shows that with syntax-enhanced representation, our model is better at dealing with lengthy questions compared with baseline.

\begin{figure*}[htb]
	\centering
	\includegraphics[width=0.85\textwidth]{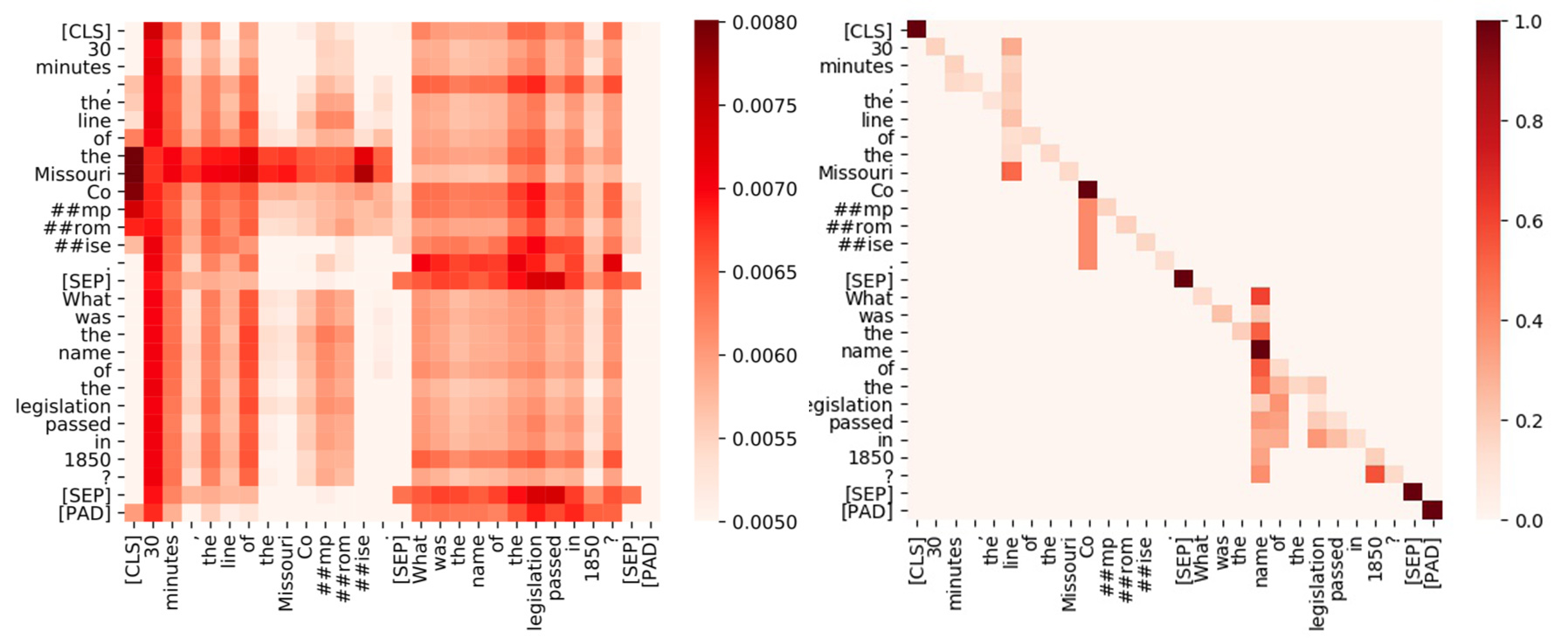}
	\\\begin{flushleft}
		\scriptsize{\emph{
				\textbf{Passage (extract)}:...30 minutes, the line of the Missouri Compromise... \textbf{Question}:What was the name of the legislation passed in 1850?  \textbf{Answer}:the Missouri Compromise} }
	\end{flushleft}
	\caption{\label{fig:vis-att} Visualization of the vanilla BERT attention (left) and syntax-guided self-attention (right). Weights of attention are selected from first head of the last attention layer. For the syntax-guided self-attention, the columns with weights represent the SDOI for each word in the row. For example, the SDOI of \textit{passed} contains \textit{\{name, of, legislation, passed\}}. Weights are normalized by SoftMax for each row. }
\end{figure*}

\subsection{Visualization}
To have an insight that how syntax-guided attention works, we draw attention distributions of the vanilla attention of the last layer of BERT and our proposed syntax-guided self-attention\footnote{Since special symbols such as [PAD] and [CLS] are not considered in the dependency parsing tree, we confine the SDOI of these tokens to themselves. So these special tokens will have value of 1 as weights over themselves in syntax-guided self-attention and we will mask these weights in the following aggregation layer.}, as shown in Figure \ref{fig:vis-att}. With the guidance of syntax, the keywords \emph{name}, \emph{legislation} and \emph{1850} in the question are highlighted, and \emph{(the)} \emph{Missouri}, and \emph{Compromise} in the passage are also paid great attention, which is exactly the right answer. The visualization verifies that benefiting from syntax-guided attention layer, our model is effective at selecting the vital parts, guiding the downstream layer to collect more relevant pieces to make predictions.

\subsection{Dual Context Mechanism Evaluation}
In SG-Net, we integrate the representations from syntax-guided attention layer and the vanilla self-attention layer in dual context layer. To unveil the contribution of each potential component, we conduct comparisons on the baseline with:

\begin{enumerate}
    \item \emph{Vanilla attention only} that adds an extra vanilla BERT attention layer after the BERT output.
    \item \emph{Syntax-guided attention only} that adds an extra syntax-guided layer after the BERT output.
    \item \emph{Dual contextual attention} that is finally adopted in SG-Net as described in Section \ref{SG_ATT}.
\end{enumerate}

\begin{table}
	\centering
	
	\begin{tabular}{l c c}
		\hline
		
		\hline
		\textbf{Model} & \textbf{EM} & \textbf{F1}\\
		\hline
		baseline & 84.1 & 86.8 \\
		+ Vanilla attention only & 84.2 &  86.9 \\
		+ Syntax-guided attention only & 84.4 & 87.2  \\
		+ Dual contextual attention & \textbf{85.1} & \textbf{87.9} \\
		\hdashline
		Concatenation &   84.5  & 87.6 \\	
		Bi-attention & 84.9 & 87.8\\
		\hline
	\end{tabular}
	
	\caption{\label{tab:ablation} Ablation study on potential components and aggregation methods on SQuAD 2.0 dev set.}
\end{table}

Table \ref{tab:ablation} shows the results. We observe that dual contextual attention yields the best performance. Adding extra vanilla attention gives no advance, indicating that introducing more parameters would not promote the strong baseline. It is reasonable that syntax-guided attention only is also trivial since it only considers the syntax related parts when calculating the attention, which is complementary to traditional attention mechanism with noisy but more diverse information and finally motivates the design of dual contextual layer. 

Actually, there are other operations for merging representations in dual context layer besides the weighted dual aggregation, such as \emph{concatenation} and \emph{Bi-attention} \cite{Seo2016Bidirectional}, which are also involved in our comparison, and our experiments show that using dual contextual attention produces the best result.

\section{Conclusion}
This paper presents a novel syntax-guided framework for enhancing strong Transformer-based encoders. We explore to adopt syntax to guide the text modeling by incorporating syntactic constraints into attention mechanism for better linguistically motivated word representations. Thus, we adopt a dual contextual architecture called syntax-guided network (SG-Net) which fuses both the original SAN representations and syntax-guided SAN representations. Taking pre-trained BERT as our Transformer encoder implementation, experiments on two major machine reading comprehension benchmarks involving span-based answer extraction (SQuAD 2.0) and multi-choice inference (RACE) show that our model can yield new state-of-the-art or comparative results in both extremely challenging tasks. This work empirically discloses the effectiveness of syntactic structural information for text modeling. The proposed attention mechanism also verifies the practicability of using linguistic information to guide attention learning and can be easily adapted with other tree-structured annotations.

\bibliography{sgnet}
\bibliographystyle{aaai}

\end{document}